\pdfoutput=1
\documentclass[conference]{IEEEtran}
\IEEEoverridecommandlockouts
\usepackage{cite}
\usepackage{amsmath,amssymb,amsfonts}
\usepackage{algorithmic}
\usepackage[linesnumbered,lined,ruled,commentsnumbered]{algorithm2e}
\usepackage{graphicx}
\usepackage{textcomp}
\usepackage{xcolor}
\usepackage{url}
\def\BibTeX{{\rm B\kern-.05em{\sc i\kern-.025em b}\kern-.08em
    T\kern-.1667em\lower.7ex\hbox{E}\kern-.125emX}}
    
\usepackage{xspace}    
\newcommand{\RebalanceStream}{\textsc{RebalanceStream}\xspace}    
\newcommand{\RebalanceStreamPlus}{\textsc{RebalanceStream+}\xspace}
\newcommand{\Base}{\textsc{Base}\xspace}
\newcommand{\ADWIN}{\textsc{Adwin}\xspace} 
\newcommand{\SMOTE}{\textsc{Smote}\xspace} 
\newcommand{\SWT}{\textsc{Swt}\xspace}

\begin{document}

\title{Incremental Rebalancing Learning \\on Evolving Data Streams\\
}

\author{
\IEEEauthorblockN{Alessio Bernardo}
\IEEEauthorblockA{\textit{DEIB - Politecnico di Milano} \\
Milano, Italy \\
alessio.bernardo@polimi.it}
\and
\IEEEauthorblockN{Emanuele Della Valle}
\IEEEauthorblockA{\textit{DEIB - Politecnico di Milano} \\
Milano, Italy \\
emanuele.dellavalle@polimi.it}
\and
\IEEEauthorblockN{Albert Bifet}
\IEEEauthorblockA{\textit{University of Waikato, New Zealand}\\
\textit{LTCI, T\'el\'ecom ParisTech, France}\\
abifet@waikato.ac.nz}
}

\maketitle

\begin{abstract}
Nowadays, every device connected to the Internet generates an ever-growing stream of data (formally, unbounded). Machine Learning on unbounded data streams is a grand challenge due to its resource constraints. In fact, standard machine learning techniques are not able to deal with data whose statistics is subject to gradual or sudden changes without any warning. Massive Online Analysis (MOA) is the collective name, as well as a software library, for new learners that are able to manage data streams. In this paper, we present a research study on streaming rebalancing. Indeed, data streams can be imbalanced as static data, but there is not a method to rebalance them incrementally, one element at a time. For this reason we propose a new streaming approach able to rebalance data streams online. Our new methodology is evaluated against some synthetically generated datasets using prequential evaluation in order to demonstrate that it outperforms the existing approaches.
\end{abstract}


\section{Introduction}
In the last few years, many machine learning techniques have been developed in order to classify data, such as decision tree, artificial neural network or support vector machine. They all work with the entire training set available and in most of cases it has to be saved in main memory. However, nowadays, every device connected to the Internet generates an ever growing stream of data (formally unbounded). Machine Learning on unbounded data streams is a grand challenge. In fact, standard techniques are not able to continuously incorporate new data. Consider, for example, a system predicting future free parking places in a silo based on the actual number of free parking places and on the number of people in the near square. Some Internet of Thing (IOT) sensors, every few seconds, register a new sample and send it right to a server. At server side, there must be an algorithm able to train a model with only a single sample at a time. Moreover, the statistics of such data are subject to gradual or sudden changes without any warning. This phenomena is known as concept drift \cite{c1} and requires an algorithm that can adjust quickly to hanging conditions \cite{c2}. Massive Online Analysis (MOA) is a library that contains new learners that are able to manage concept drift\footnote{\url{https://moa.cms.waikato.ac.nz/}}.

Concept drift is also related to the imbalancement of the data stream. In a static approach, working with an imbalanced dataset can worsen the prediction performances. To resolve the problem, there are some techniques able to rebalance a dataset before using it to train a model. In an adaptive approach, there is not a technique able to do it without having the entire dataset available. In this study, we propose \RebalanceStream, a novel method able to rebalance data streams one element at a time. We also show that it outperforms state of the art evaluating it against some synthetically generated datasets using prequential evaluation.

The remainder of this paper is organized as follows. In Section 2, we present some existing techniques on top of which we build our contribution. In Section 3, we motivate and describe the method proposed. Section 4 describes the datasets used in our experiments and shows the evaluation results. Section 5 discusses the conclusions we reached based on these experiments and outlines directions for future research.

\section{RELATED WORK}
There are some new techniques able to work with data streams. The VFDT algorithm of Domingos and Hulten \cite{c3} builds incrementally a decision tree, using a small subset of examples to determine which attribute to use to split at a given node. They employ Hoeffding bounds to show that the resulting tree can be made arbitrarily similar to one that would be built having all the data at hand. Another approach is called CVFDT and is presented in \cite{c4}. It works by keeping its model consistent with respect to a sliding window of data from the data stream, and creating and replacing alternate decision subtrees when it detects that the distribution of data is changing at a node. The drawback is that it does not automatically detect the optimal window size like \ADWIN does. Bifet and Gavald\'a \cite{c2} develop other approaches based on VFDT algorithm, such as HWT-\ADWIN and HAT that use \ADWIN \cite{c5} as a change detector. It keeps a variable-length window of recently seen items and it is able to automatically detect and adapt its window to the current rate of change. A user of \ADWIN has only to decide how to measure error (e.g., using accuracy). \ADWIN, in~\eqref{eq:1}, uses a threshold called $delta$ in order to automatically configure the error with two levels, named warning and change level. The warning level is identified using $delta\times10$, while the change level is identified using $delta$. Since $delta$ appears to the denominator, using $delta\times10$ will produce a lower value than using $delta$. So the warning level will occur before the change one. $n$ is the width of the window in that moment. \ADWIN monitors the error over the data in the window. If the error becomes greater than a warning level, \ADWIN assumes that a concept drift starts to occur and it starts collecting new samples in a new window, too. If the error becomes greater than the change level, \ADWIN assumes that a concept occurred and it substitutes the old window with the new one. 

\begin{equation} \label{eq:1}
levelError = \ln(\frac{2\times\ln{n}}{delta})
\end{equation}

HWT-\ADWIN is a new Hoeffding Window Tree, while HAT evolves from HWT and replaces, at each node, the original counter with an \ADWIN instance. \ADWIN instances are also change detectors, so they notice when a change in the statistics at that node is detected, which can also be a possible concept change. Gomes et al. \cite{c6} propose a technique called Adaptive Random Forest (ARF). It is the adaptation of Random Forest (RF) algorithm \cite{c7} to work with streaming data. RF combines multiple hypothesis from multiple decision trees in order to form a final better one. It grows a lot of decision trees at training time and gives in output the most popular class. To avoid overfitting, the idea is to combine bootstrap aggregating and feature bagging together. Bootstrap aggregating repeatedly selects a random sample with replacement of the training set $b$ times and grows trees with these new samples, i.e. one tree for each sample. Feature bagging uses a random subset of the features to make the split on each tree. ARF uses the Online Bagging procedure instead of Bootstrap aggregating. Online Bagging sends $k$ copies of each new examples to update each model, where $k$ is a suitable Poisson random variable. Moreover ARF also uses \ADWIN to detect warnings and create “background” trees that are trained along the ensemble without influencing the ensemble predictions. If a drift is detected from the tree originating the warning signal, the original tree is then replaced by its respective background tree. Finally, we would like to mention \SWT \cite{c8} by Biffet et al. This is  an algorithm that uses a meta strategy to build meta instances by increasing the original input attributes adding attributes with the values of the most recent class labels from previous samples. \newline
All these adaptive methods use the prequential evaluation approach \cite{c9} in order to test their models. Instead of using a static batch of data to test the model (testing set), this approach firstly tests the model with the new incoming sample and then uses it to train the model. In this way, the model is always tested on data it has never seen before. This approach is introduced due to the lack of a static batch to use as testing set. For every new sample, after prequential evaluation, the performance result is saved in a confusion matrix that will be used to compute the K-statistic score. \newline
K-statistic \cite{c9} is a new performance metric able to take the stream evolution into account. It states if a method is a good classifier, respect to a chance classifier (roughly speaking, a random guesser), for the phenomena that it is trying to predict. Equation~\eqref{eq:2} shows how to calculate it. The quantity $p_0$ is the classifier’s prequential accuracy while $p_c$ is the probability that a chance classifier, the one that assigns the same number of examples to each class as the classifier under consideration, makes a correct prediction. If the classifier is always correct then $k = 1$. If its predictions are correct as often as those of a chance classifier then $k = 0$.

\begin{equation} \label{eq:2}
k = \frac{(p_0 - p_c)}{(1 - p_c)}
\end{equation}

Regarding the imbalancement problem, it may be considered a serious problem for model learning: in fact, a learner can analyze the data and cleverly decide that the best thing to do is to always predict the majority class without performing any analysis of the features. In the static settings there are two approaches to rebalance a dataset: under-sampling the majority class removing random records or oversampling the minority class replicating random existing records. One of the most powerful techniques is \SMOTE \cite{c10}. It over-samples the minority class at a certain percentage by creating synthetic samples. For each minority class sample, \SMOTE finds its $k$ nearest neighbours from the minority class samples, it randomly chooses one from them and uses it to create synthetic samples. 

Since data streams can evolve over time, the number of samples for each class may also change. It may be that the majority of samples always have the same label. So streams can be imbalanced, too. The techniques previously described and also all the other existing ones work with a static batch. In fact they must know the number of elements in the majority and minority class in order to set a percentage of rebalancement. In case of evolving batch, this is impossible to know. Data arrive continuously and they are unbounded, too.

\section{PROPOSED METHOD}
It is well known \cite{c10} that models using a balanced static batch can have higher performances respect to those that use an imbalanced dataset. It can be interesting to know if this aspect is valid for streaming methods, too. Our proposal is a meta strategy, as \SWT, called \RebalanceStream able to rebalance a stream and train a model with it. It is represented as pseudo-code in Alg.~\ref{alg1}. The full code is available online on a GitHub repository\footnote{\url{https://github.com/alessiobernardo/RebalanceDataStream}}.

The general idea of \RebalanceStream is to use \ADWIN, as ARF does, in order to detect when there is a concept drift in the stream and be able to adapt the model under construction. When this happens, the aim is to use \SMOTE to rebalance the data arrived up to that point and to use the rebalanced data to train other models. The best trained model is chosen in order to continue the execution. More specifically, before the first sample arrives, the proposed method initializes the four models used (lines 2-5). All of them use a \SWT classifier with ARF as base learner. For each new sample \textit{trainInst} that arrives, the algorithm does the prequential evaluation, it updates the \textit{confusionMatrix} and the \ADWIN estimator. Then, it trains the model called \textit{learner} and it saves \textit{trainInst} in a \textit{batch} (lines 7-11). When \textit{\ADWIN} detects a warning, the algorithm starts collecting \textit{trainInst} also in a new batch called \textit{resetBatch} (lines 15-16). If \textit{trainInst} is the \textit{n}th instance or its multiple, it uses the actual state of the \textit{confusionMatrix} to calculate the prequential evaluation k-statistic and it saves it in the \textit{kp} list (line 18-21). When \textit{\ADWIN} detects a change, it uses the \textit{confusionMatrix} to calculate the \textit{kStatLearner} and it trains three other models:

\begin{itemize}
\item
\emph{LearnerBal}: it applies \SMOTE on the batch and the method uses it to train the \textit{learnerBal} model, in the same way as before. It returns the \textit{kStatBal}. The pseudocode is shown in Alg.~\ref{alg2}.

\item
\emph{LearnerReset}: it uses the \textit{resetBatch} to train the \textit{learnerReset} model. It returns the \textit{kStatReset}. The pseudocode is shown in Alg.~\ref{alg3}.

\item
\emph{LearnerResetBal}: it applies \SMOTE on the \textit{resetBatch} and the method uses it to train the \textit{learnerResetBal} model. It returns the \textit{kStatResetBal}. The pseudocode is shown in Alg.~\ref{alg4}.

\end{itemize}

\begin{algorithm}[tbp]
\SetAlgoLined
\SetKwFunction{FMain}{run}
\SetKwFunction{trainLearnerBal}{trainLearnerBal}
\SetKwFunction{trainLearnerReset}{trainLearnerReset}
\SetKwFunction{trainLearnerResetBal}{trainLearnerResetBal}
\SetKwProg{Fn}{Function}{:}{}
\Fn{\FMain{$train$}}{
    Initialize $learner$\;
    Initialize $learnerBal$\;
    Initialize $learnerReset$\;
    Initialize $learnerResetBal$\;
    \ForEach{$trainInst \in train$}{
       $trainInst$ prequential evaluation\;
       Update the $confusionMatrix$\;
       Update $\ADWIN$\;
       Train $learner$ with $trainInst$\;
       Add $trainInst$ to $batch$\;
       \If{$\ADWIN$ detects a $warning$}{
         $w \gets true$\;
       }
       \If{$w$ == $true$}{
       Add $trainInst$ to $resetBatch$\;
       }
       \If{$trainInst$ is the $n$th instance or its multiple}{
       $kStat \gets$ k-statistics $confusionMatrix$\; 
       Add $kStat$ to $kp$\;
       }
       \If{$\ADWIN$ detects a $change$}{
        $kStatLearner \gets$ k-statistics $confusionMatrix$\;
        $kStatBal \gets$ \trainLearnerBal{$batch$}\;
        $kStatReset \gets$ \trainLearnerReset{$resetBatch$}\;
        $kStatResetBal \gets$ \trainLearnerResetBal{$resetBatch$}\;
        $max \gets$ max between all kStat\;
        $learner \gets$ model having $max$\;
        $confusionMatrix \gets$ confusionMatrix of model having $max$\;
        Reset other models, $batch$ and $resetBatch$\;
        $w \gets false$\;
        }
    }
    \KwRet $kp$\;
}
\textbf{End Function}
\caption{\RebalanceStream algorithm pseudocode}
\label{alg1}
\end{algorithm}

Finally, Alg.~\ref{alg1} at line 27 chooses the highest k-statistic value among the four k-statistics previously calculated and finds the associated model. It swaps the model called \textit{learner} with the best model found before and the \textit{confusionMatrix} with the one corresponding to the model found (lines 28-29). At the end, it resets all the other models and data  structures (lines 30-31). The model called \textit{learner} and the \textit{confusionMatrix} will be used to continue with the new samples. 

\begin{algorithm}[h]
\DontPrintSemicolon
\SetAlgoLined
\SetKwFunction{FMain}{trainLearnerBal}
\SetKwProg{Fn}{Function}{:}{}
\Fn{\FMain{$batch$}}{
Use $\SMOTE$ to rebalance $batch$\;
\ForEach{$inst \in batch$}{
Do prequential evaluation with $inst$ on $learnerBal$\;
Update the $confusionMatrixBal$\;
Train $learnerBal$ with $inst$\;
}
$kStatBal \gets$ k-statistics $confusionMatrixBal$\;
\KwRet $kStatBal$
}
\textbf{End Function}
\caption{learnerBal training pseudocode}
\label{alg2}
\end{algorithm}

\begin{algorithm}[h]
\DontPrintSemicolon
\SetAlgoLined
\SetKwFunction{FMain}{trainLearnerReset}
\SetKwProg{Fn}{Function}{:}{}
\Fn{\FMain{$resetBatch$}}{
\ForEach{$inst \in resetBatch$}{
Do prequential evaluation with $inst$ on $learnerReset$\;
Update the $confusionMatrixReset$\;
Train $learnerReset$ with $inst$\;
}
$kStatReset \gets$ k-statistics $confusionMatrixReset$\;
\KwRet $kStatReset$\;
}
\textbf{End Function}
\caption{learnerReset training pseudocode}
\label{alg3}
\end{algorithm}

\begin{algorithm}[h]
\DontPrintSemicolon
\SetAlgoLined
\SetKwFunction{FMain}{trainLearnerResetBal}
\SetKwProg{Fn}{Function}{:}{}
\Fn{\FMain{$resetBatch$}}{
Use $\SMOTE$ to rebalance $resetBatch$\;
\ForEach{$inst \in resetBatch$}{
Do prequential evaluation with $inst$ on $learnerResetBal$\;
Update the $confusionMatrixResetBal$\;
Train $learnerResetBal$ with $inst$\;
}
$kStatResetBal \gets$ k-statistics $confusionMatrixReset$\;
\KwRet $kStatResetBal$
}
\textbf{End Function}
\caption{learnerResetBal training pseudocode}
\label{alg4}
\end{algorithm}

\begin{algorithm}[h]
\SetAlgoLined
\SetKwFunction{FMain}{run}
\SetKwProg{Fn}{Function}{:}{}
\Fn{\FMain{$train$}}{
    Initialize $learner$\;
    \ForEach{$trainInst \in train$}{
       $trainInst$ prequential evaluation\;
       Update the $confusionMatrix$\;
       Train $learner$ with $trainInst$\;
       \If{$trainInst$ is the $n$th instance or its multiple}{
       $kStat \gets$ k-statistics $confusionMatrix$\; 
       Add $kStat$ to $kp$\;
       }
    }
    \KwRet $kp$\;
}
\textbf{End Function}
\caption{\Base algorithm pseudocode}
\label{alg5}
\end{algorithm}

\begin{figure*}[tp]
\centerline{\includegraphics[width=18cm,height=6cm]{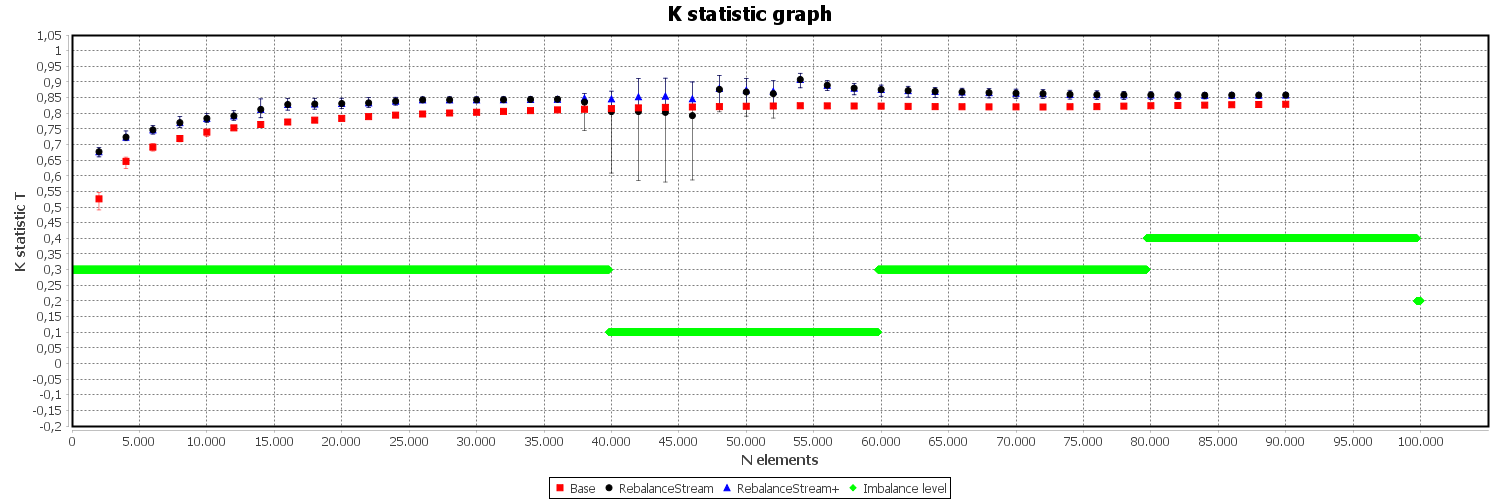}}
\caption{Line chart that compares \Base, \RebalanceStream and \RebalanceStreamPlus algorithms with mean = 20000 and variance = 50.}
\label{fig1}
\end{figure*}

The proposed algorithm is compared to its base version, called \Base, shown in Alg.~\ref{alg5}. Also in this case, the model called \textit{learner} uses a \SWT classifier with ARF as base learner. For each new sample \textit{trainInst} that arrives, the \Base algorithm does the prequential evaluation and updates the \textit{confusionMatrix}. Then, it trains the \textit{learner} (lines 4-6). If \textit{trainInst} is the \textit{n}th instance or its multiple, it uses the actual state of the \textit{confusionMatrix} to calculate the prequential evaluation k-statistic and it saves it in the \textit{kp} list (line 7-10). When all the \textit{trainInst} samples are over, it returns the \textit{kp} list.

We propose another approach called \RebalanceStreamPlus, too. We run in parallel the \RebalanceStream and \Base algorithm. When the dataset imbalance level is high, and the \RebalanceStream results are worse than the \Base ones, we use the \Base algorithm results. In this way, we create a set of results shown in the next section. 

\section{EXPERIMENTAL RESULTS}
The datasets used in the experiments are a synthetically generated through a RandomRBFGeneratorDrift in MOA\footnote{\url{https://moa.cms.waikato.ac.nz/}}. We generate five different datasets. Each of them uses a different seed value. The values used are 3, 4, 5, 6 and 7. Moreover, each dataset is composed by 100,000 samples and uses fifty centroids, a speed change of 0.0000001, ten attributes and two classes. In order to change the imbalancement level, the class ratio is randomly chosen from values between (0.6;0.4) and (0.9;0.1) and it is changed every some random numbers of rows. The number of rows among changes are randomly sampled from a Gaussian model with a certain mean $\mu$ and variance $\sigma^2$. In order to do some tests, we use different combinations of $\mu$ and $\sigma^2$. The $\mu$ values used are 20,000, 22,500, 25,000, 27,500 and 30,000. The $\sigma^2$ values used are 50, 100, 200 and 400. For these experiments, we used $n = 2000$ at line 18 of Alg.~\ref{alg1} and at line 7 of Alg.~\ref{alg5}. The green line in Fig.~\ref{fig1}, as well as all those in the small figures in Tab.~\ref{tab1}, illustrates the imbalancement level and the way it changes during an experiment. In particular, Tab.~\ref{tab1} allows seeing how the lines are longer (a given imbalancement level last longer) when the mean $\mu$ is larger.

\newpage
After the execution of \Base, \RebalanceStream and \RebalanceStreamPlus algorithms, we have five sets of results for each method. In order to aggregate and plot them, we make "vertical" summaries for each algorithm. For all the elements in the time-series of the results of each algorithm, we make the mean among the \textit{i} values of sets $s_1$, $s_2$, $s_3$, $s_4$ and $s_5$ and we find the minimum and maximum value. Finally, we have three time-series for each algorithm: the mean of k-statistic (namely, $mean$), the minimum values (namely, $min$) and the maximum values (namely, $max$). These are shown in a line chart with error bars in Fig.~\ref{fig1} as well as in all the small figures in Tab.~\ref{tab1}.

Fig.~\ref{fig1} shows the line chart having $\mu = 20000$ and $\sigma^2 = 50$. The red line represents \Base algorithm results, the black one represents \RebalanceStream algorithm results, while the blue one represents \RebalanceStreamPlus algorithm results. Moreover the green line represents the imbalance level of the five datasets. Most of the time, both the blue and the black lines are greater than the red one. It means that the results of our novel algorithms outperform the results of \Base algorithm. Nevertheless, there is a phenomenon to notice: when the minority class is 10\% of the entire data, only the \RebalanceStreamPlus algorithm results are better than the \Base ones, while the \RebalanceStream ones are worse. In some datasets, the minimum value is very low w.r.t. the mean value. The line charts of all the combinations are organized in Tab.~\ref{tab1} in order to be easily compared. The phenomenon previously described is visible in all the charts.

We also create a heatmap that allows to easily compare the results from all the $\mu$ and $\sigma^2$ combinations for a pair of algorithms. On the columns there are all the $\mu$ values, while the $\sigma^2$ values are on the rows. In our case, we have two heatmaps: one to compare \RebalanceStream and \Base algorithms and the other one to compare \RebalanceStreamPlus and \Base algorithms. Both heatmaps are created by starting from two matrix of results. For every experiment, i.e., for each combination of $\mu$ and $\sigma^2$, we create two aggregated values that sum up the overall result of the experiment, one for each couple of algorithms. This value is meant to tell at a glace if a method is better or worse than the other one. A single value is calculated from the $mean_1$ and $mean_2$ sets of results of the two algorithms to compare. For all their elements, we take the $i$ element from $mean_1$ and $mean_2$ and we calculate the difference. At the end, we make the mean of all the differences and we obtain a single value. If it is positive, it means that the first algorithm is better than the second one, otherwise the second algorithm is better than the first one. In the heatmap, if the value is positive, it is shown in green.

\begin{figure}[h]
\centerline{\includegraphics[width=0.9\columnwidth]{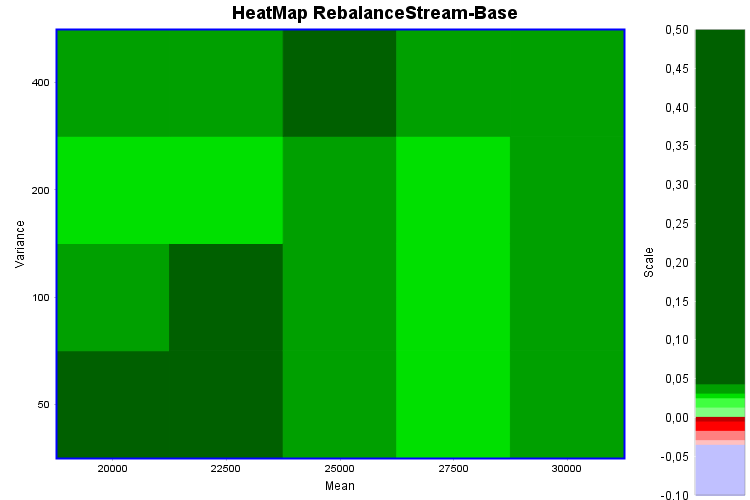}}
\caption{Heatmap that compares \RebalanceStream algorithm with \Base algorithm}
\label{fig2}
\end{figure}

Fig.~\ref{fig2} shows the heatmap that allows easily comparing \RebalanceStream and \Base algorithms. All the $\mu$ and $\sigma^2$ combinations are green, even if, in some combinations, the value is closer to zero in respect to the other ones (light green). In general, it means that the mean of the differences is positive and that the \RebalanceStream algorithm is better than the \Base one.

\section{CONCLUSIONS}
We have presented two novel approaches able to rebalance data streams one element at a time and we have evaluated them against some synthetically generated datasets using prequential evaluation. All the results empirically prove that, with prequential evaluation results, rebalancing a data stream increases the performances as in the case of static batches. Therefore, the algorithms proposed show a valid approach to rebalance an imbalance data stream. 

The next step of this research is to understand the reasons why the results get worse when the minority class level is less than 10\% of total data and to find a solution. In the long term we will focus on evaluating \RebalanceStream and \RebalanceStreamPlus algorithms against real-world data streams, with a different number of attributes and classes. Moreover, we will focus on comparing the methods proposed in terms of computing time, too.

\section*{Acknowledgment}
Some initial ideas for this paper was conceived at Dagstuhl Seminar 17441 on “Big Stream Processing Systems”.

\begin{table*}[htbp]
\caption{Line charts of prequential evaluation results of all the combinations between $\mu$ and $\sigma^2$. Red lines represent \Base algorithm, black lines represent \RebalanceStream algorithm, blue lines represent \RebalanceStreamPlus algorithm and green lines represent the dataset imbalance level.}
\begin{center}
\begin{tabular}{|c|c|c|c|c|}
\hline
& \multicolumn{4}{|c|}{\textbf{$\sigma^2$}} \\
\cline{2-5} 
\textbf{$\mu$} & \textbf{\textit{50}} & \textbf{\textit{100}} & \textbf{\textit{200}} & \textbf{\textit{400}} \\

\hline
\parbox[t]{2mm}{\rotatebox[origin=c]{90}{20,000}} & 
\parbox[c][3.6cm][c]{4cm}{\includegraphics[width=4cm,height=3.3cm]{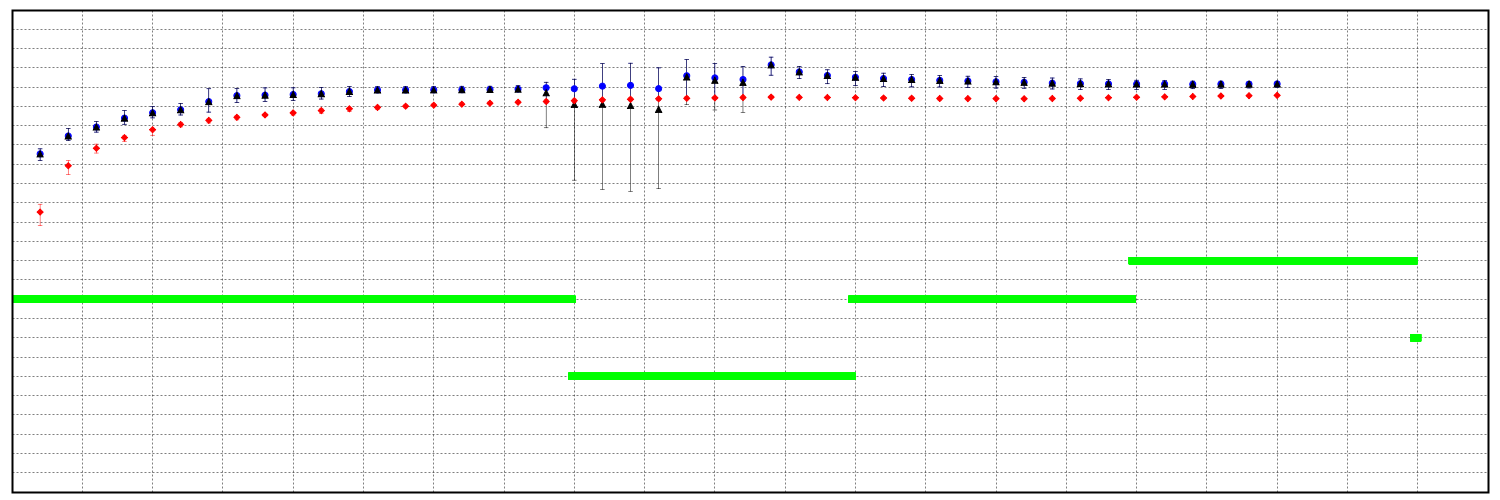}} & \parbox[c][3.6cm][c]{4cm}{\includegraphics[width=4cm,height=3.3cm]{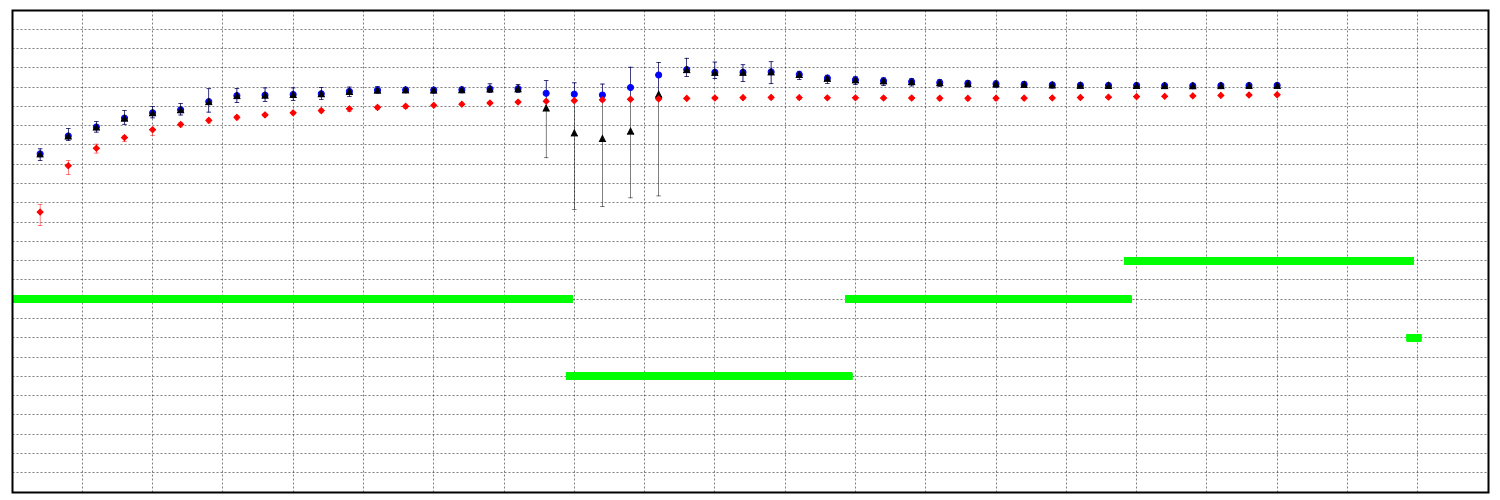}} &
\parbox[c][3.6cm][c]{4cm}{\includegraphics[width=4cm,height=3.3cm]{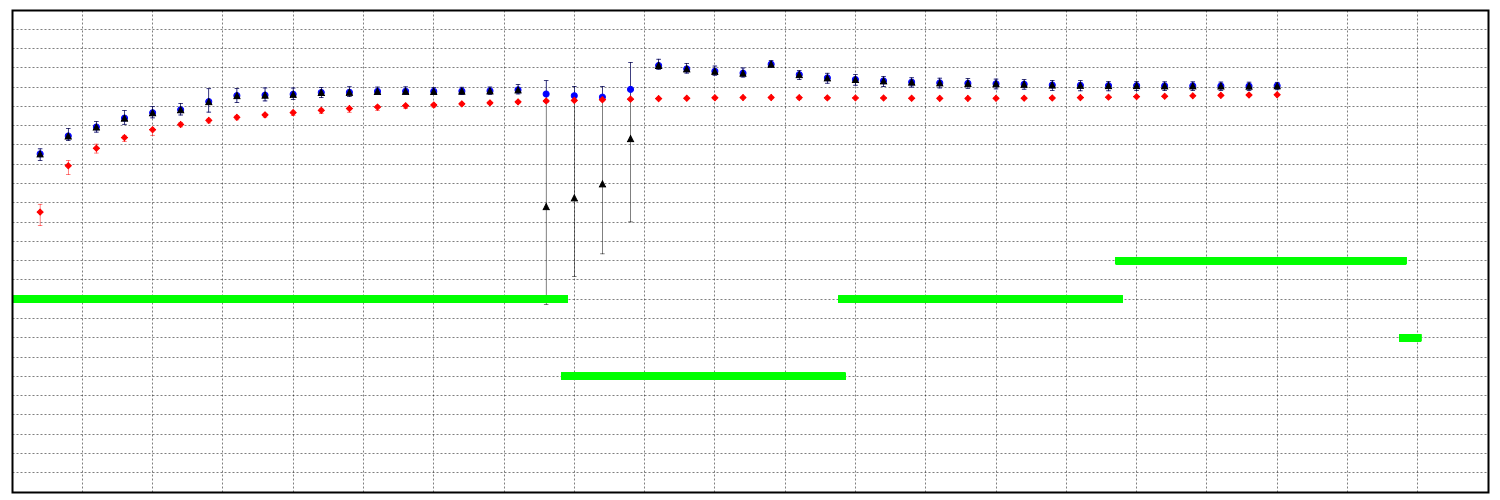}} &
\parbox[c][3.6cm][c]{4cm}{\includegraphics[width=4cm,height=3.3cm]{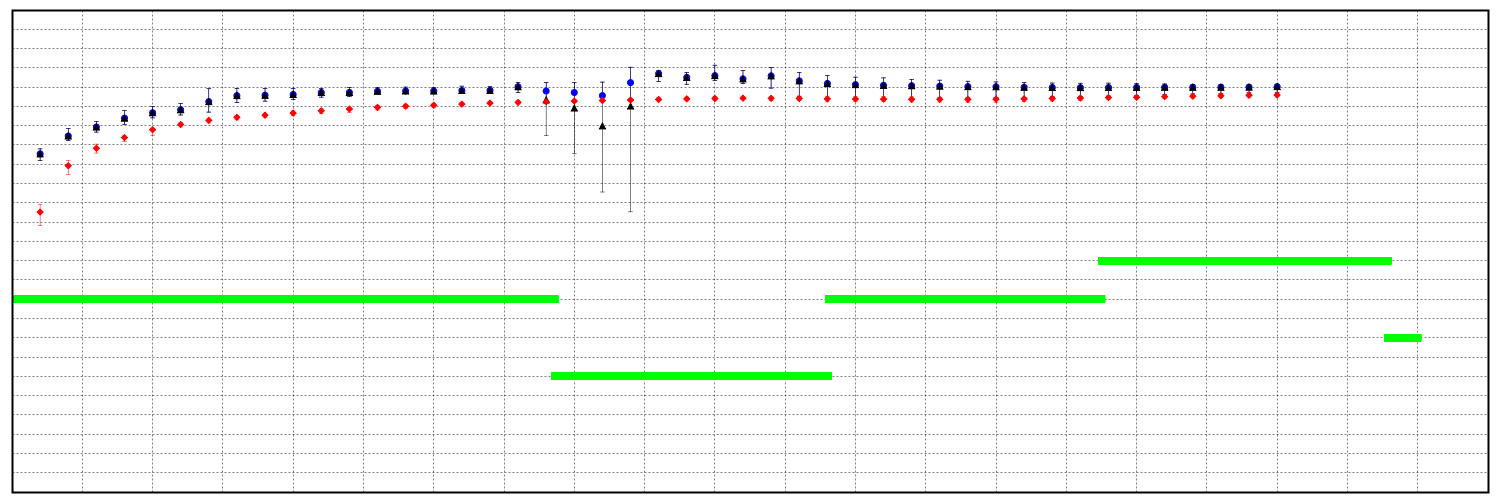}} \\
\hline

\hline
\parbox[t]{2mm}{\rotatebox[origin=c]{90}{22,500}} & 
\parbox[c][3.6cm][c]{4cm}{\includegraphics[width=4cm,height=3.3cm]{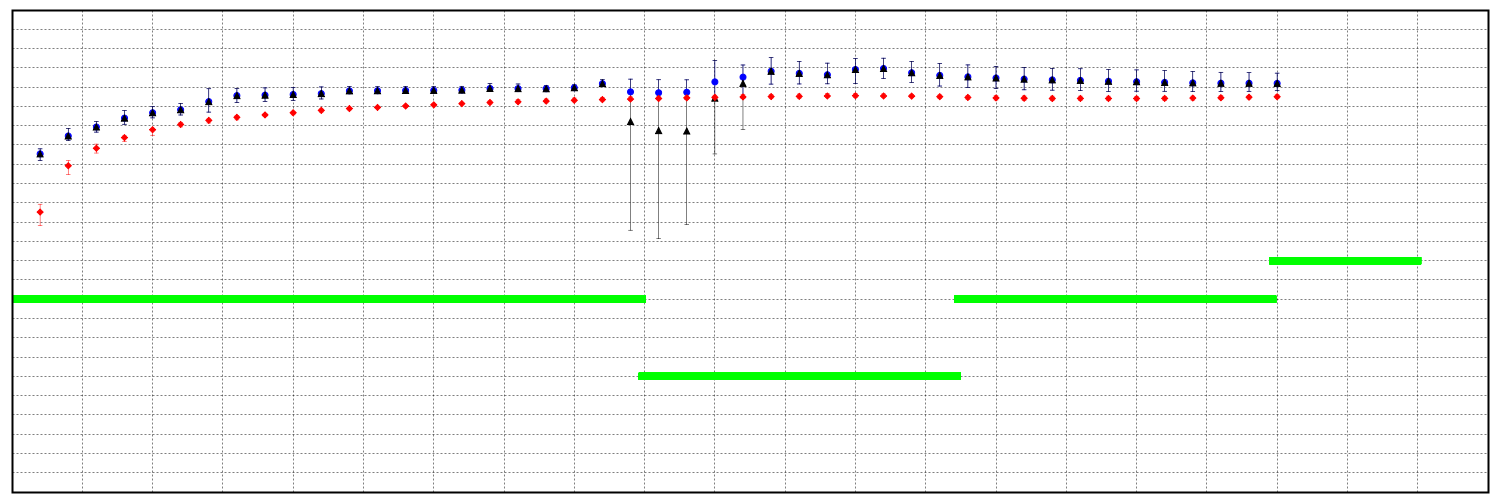}} & \parbox[c][3.6cm][c]{4cm}{\includegraphics[width=4cm,height=3.3cm]{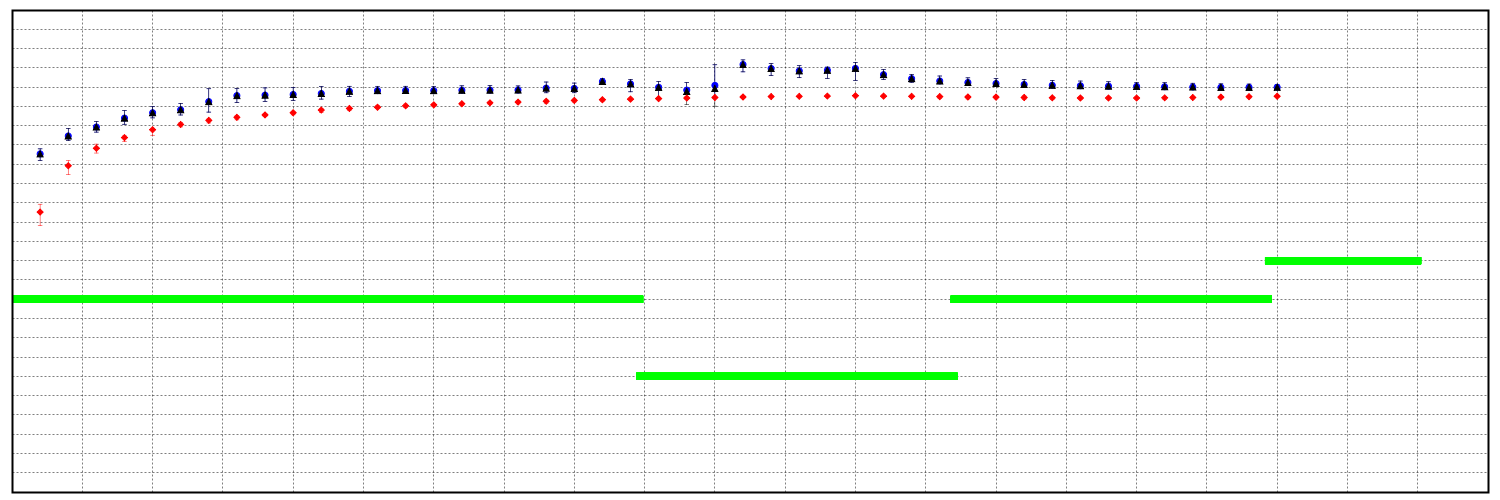}} &
\parbox[c][3.6cm][c]{4cm}{\includegraphics[width=4cm,height=3.3cm]{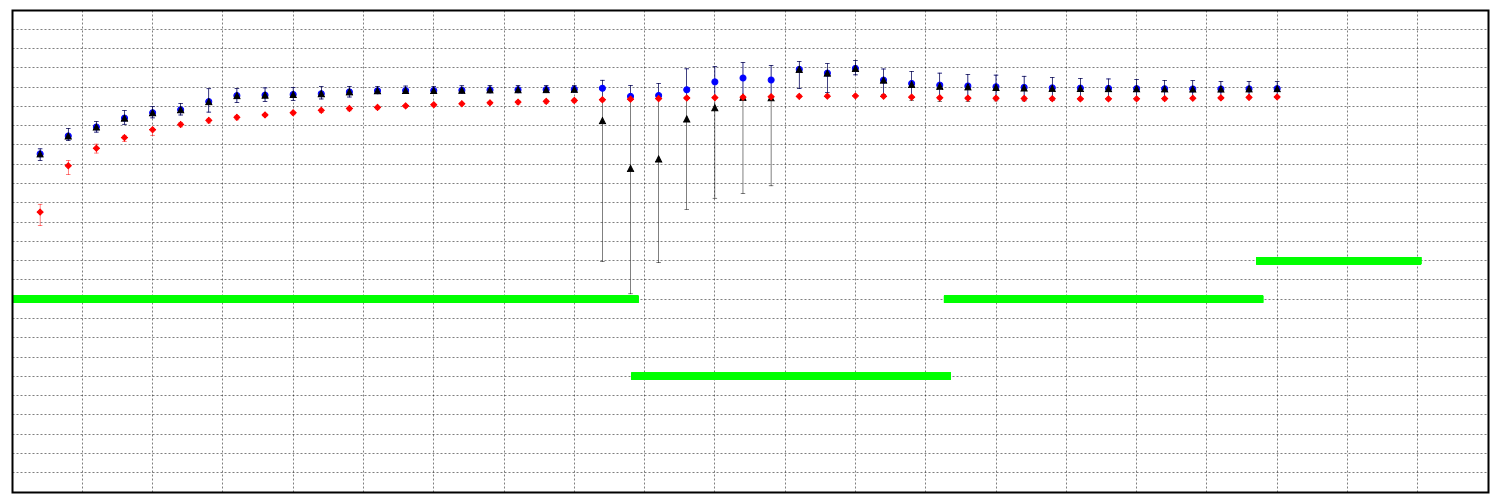}} &
\parbox[c][3.6cm][c]{4cm}{\includegraphics[width=4cm,height=3.3cm]{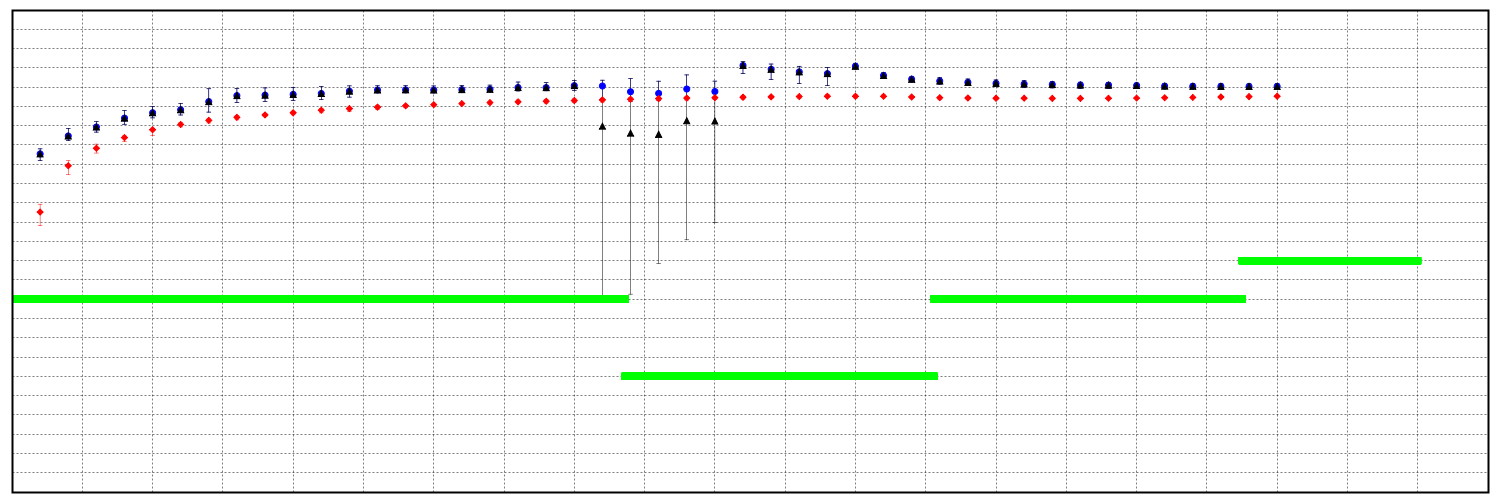}} \\
\hline

\hline
\parbox[t]{2mm}{\rotatebox[origin=c]{90}{25,000}} & 
\parbox[c][3.6cm][c]{4cm}{\includegraphics[width=4cm,height=3.3cm]{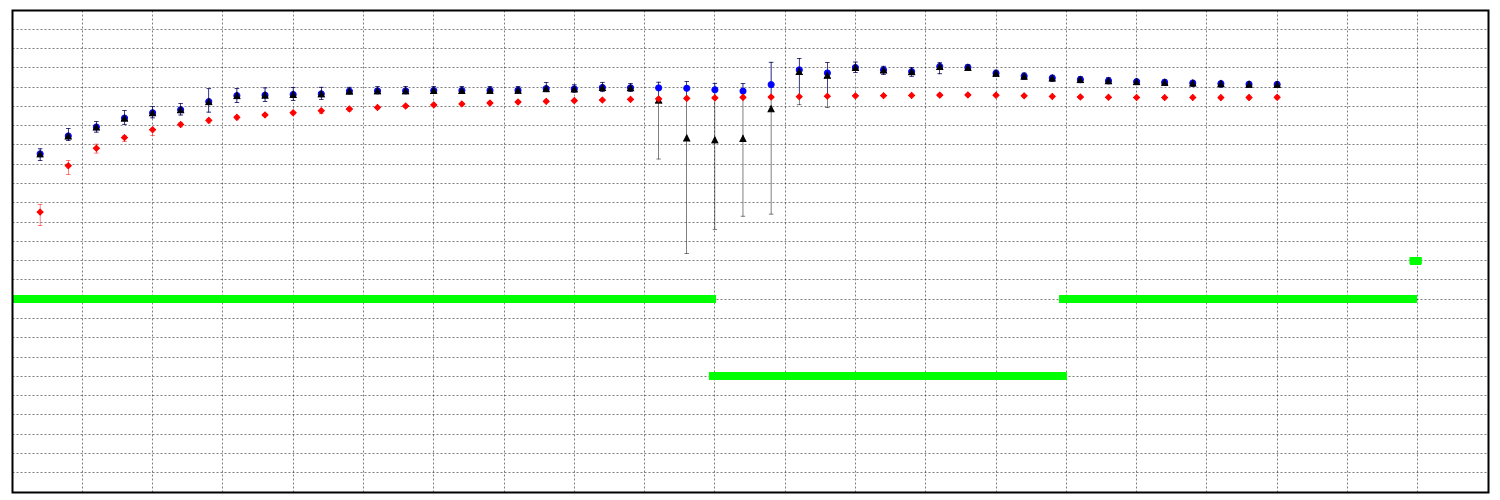}} & \parbox[c][3.6cm][c]{4cm}{\includegraphics[width=4cm,height=3.3cm]{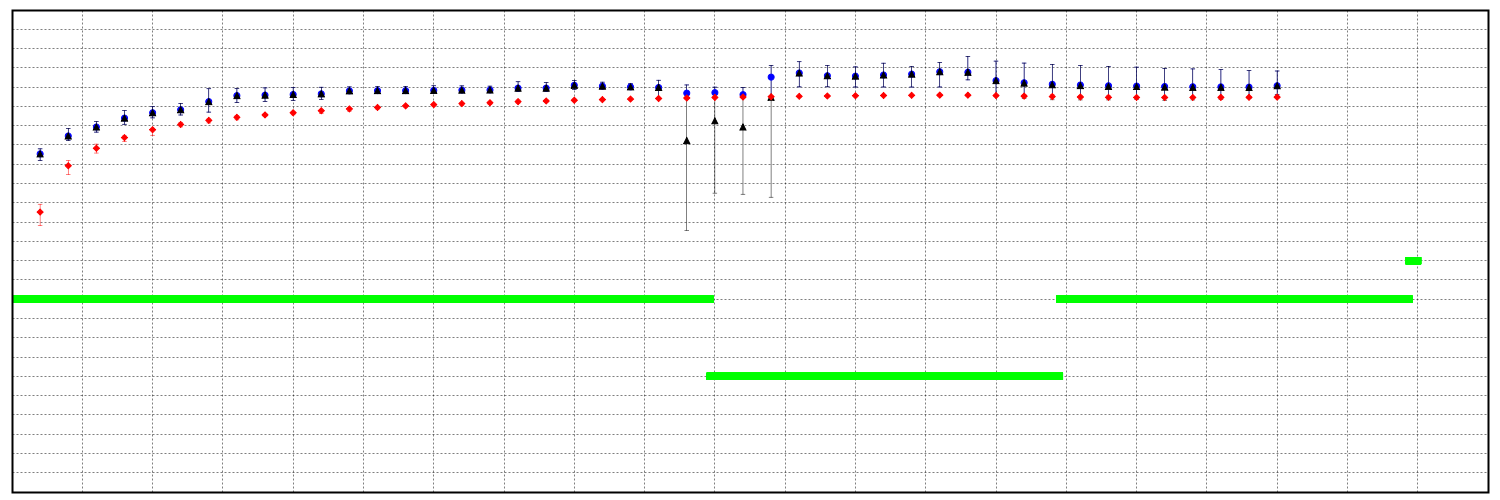}} &
\parbox[c][3.6cm][c]{4cm}{\includegraphics[width=4cm,height=3.3cm]{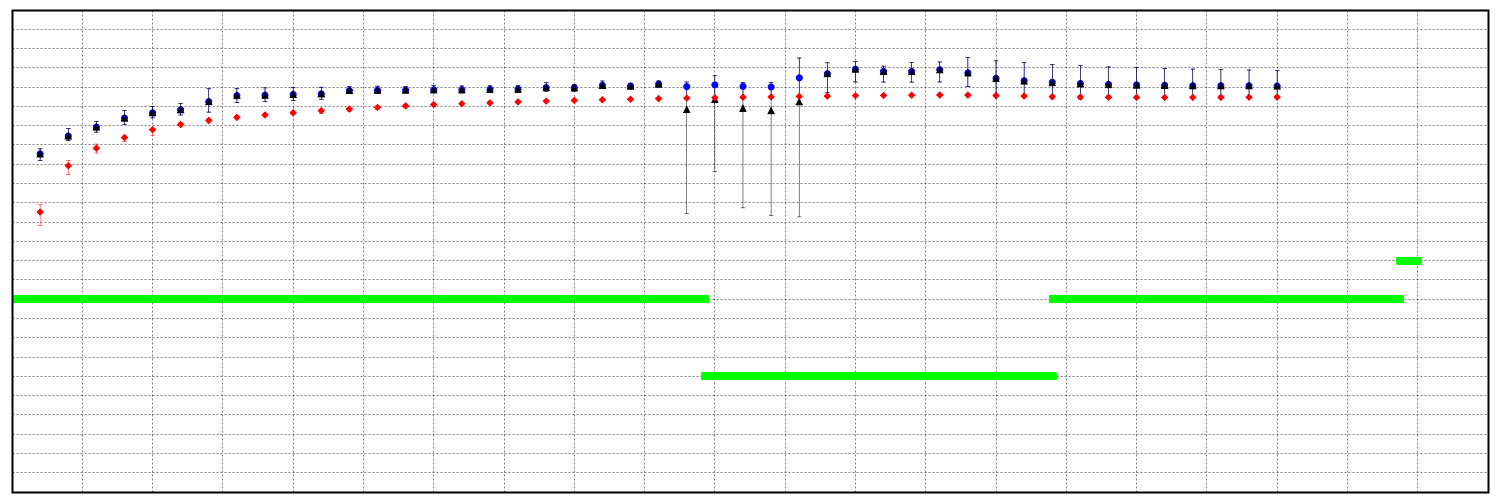}} &
\parbox[c][3.6cm][c]{4cm}{\includegraphics[width=4cm,height=3.3cm]{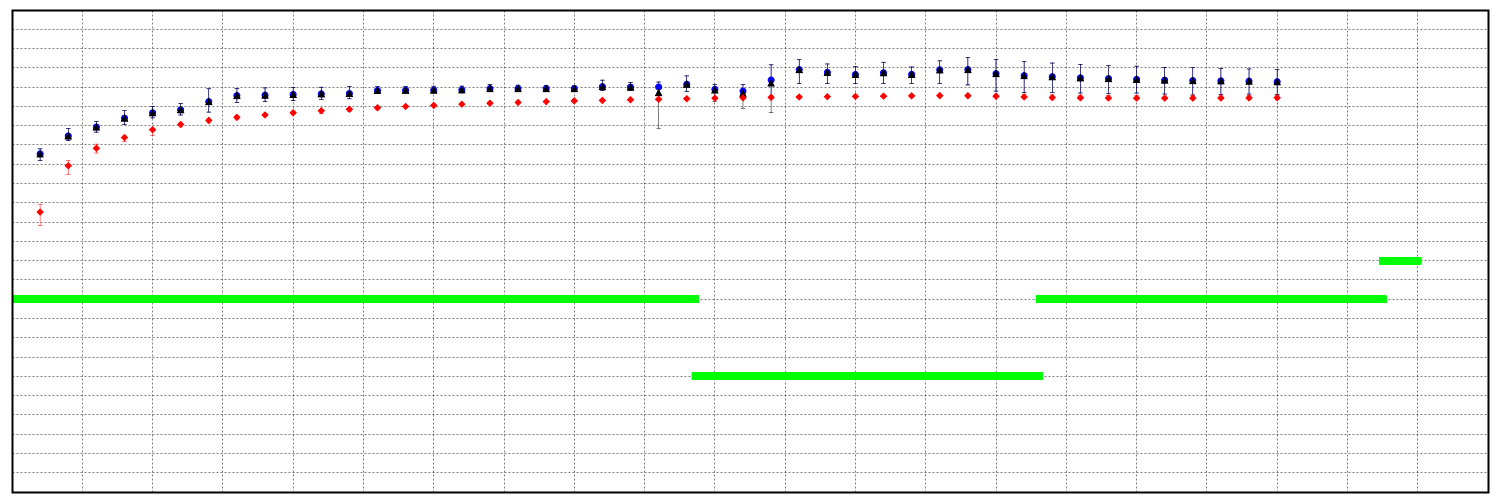}} \\
\hline

\hline
\parbox[t]{2mm}{\rotatebox[origin=c]{90}{27,500}} & 
\parbox[c][3.6cm][c]{4cm}{\includegraphics[width=4cm,height=3.3cm]{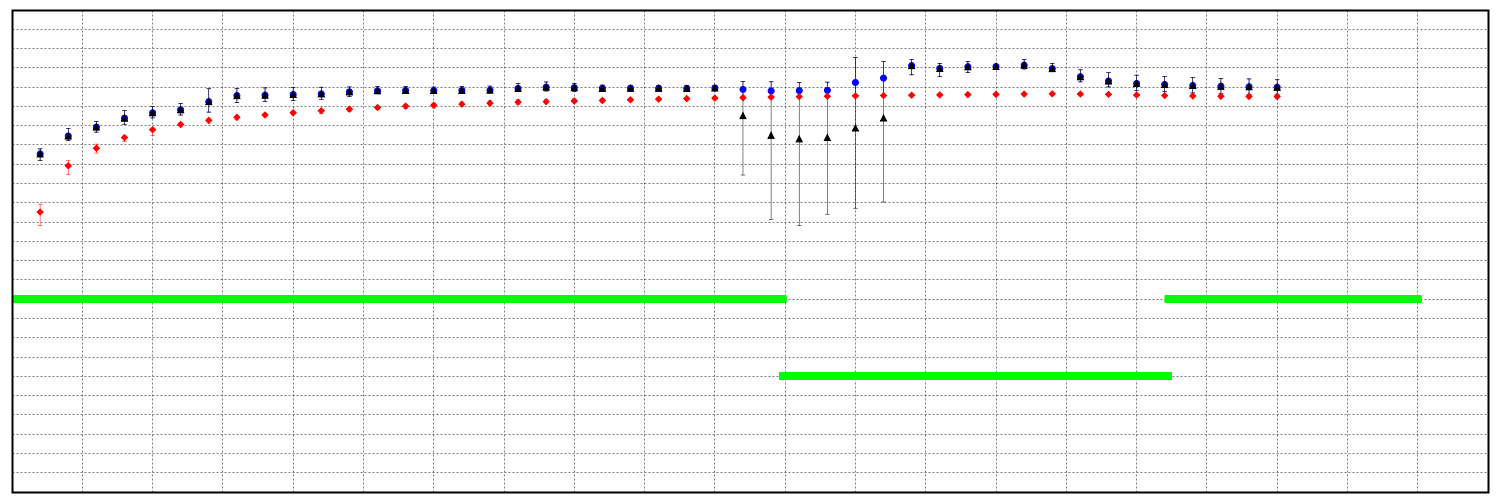}} & \parbox[c][3.6cm][c]{4cm}{\includegraphics[width=4cm,height=3.3cm]{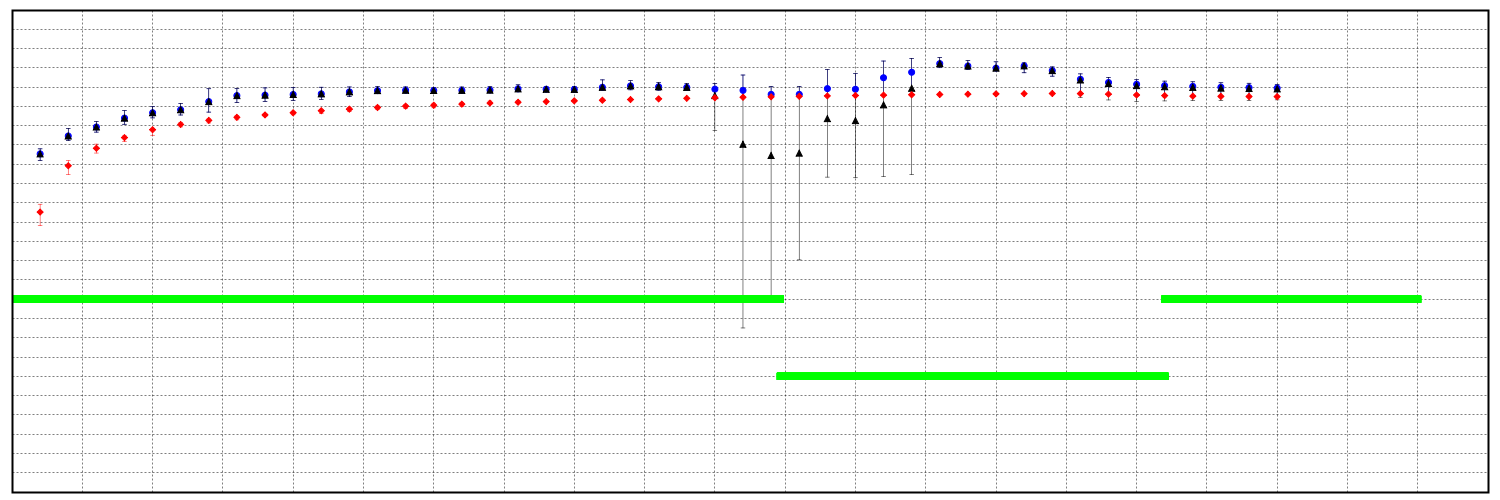}} &
\parbox[c][3.6cm][c]{4cm}{\includegraphics[width=4cm,height=3.3cm]{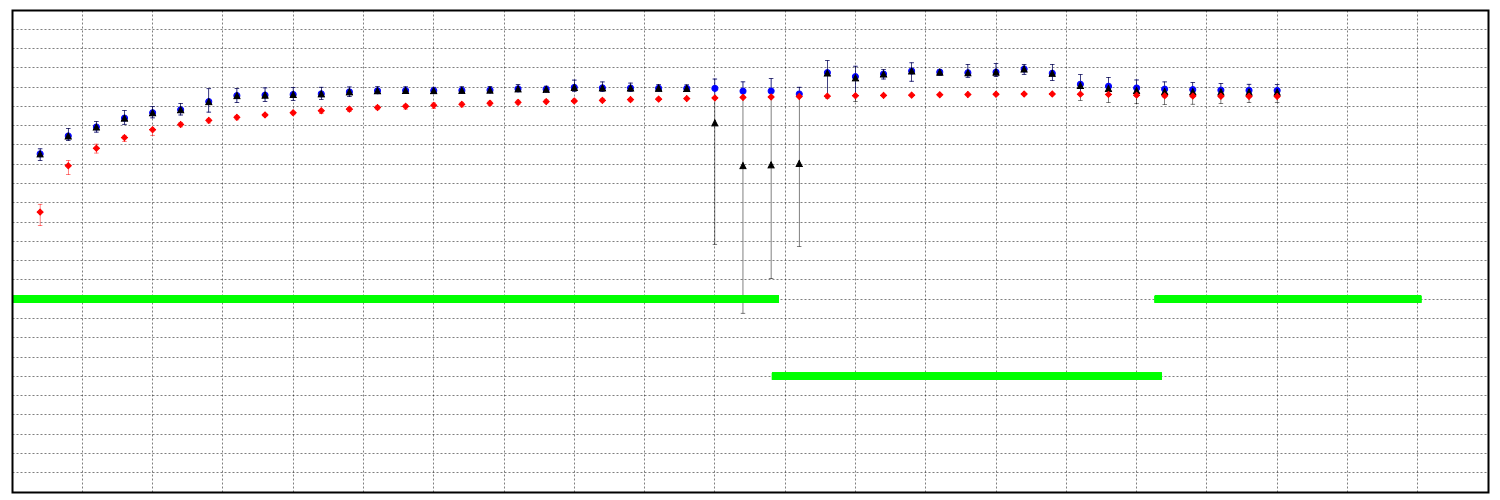}} &
\parbox[c][3.6cm][c]{4cm}{\includegraphics[width=4cm,height=3.3cm]{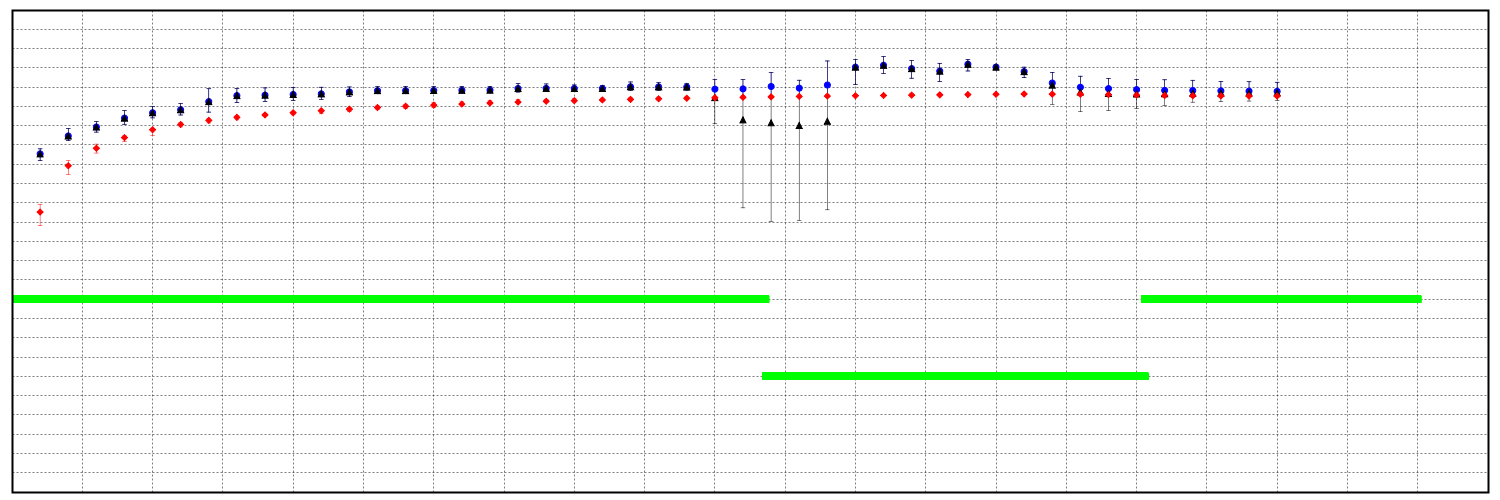}} \\
\hline

\hline
\parbox[t]{2mm}{\rotatebox[origin=c]{90}{30,000}} & 
\parbox[c][3.6cm][c]{4cm}{\includegraphics[width=4cm,height=3.3cm]{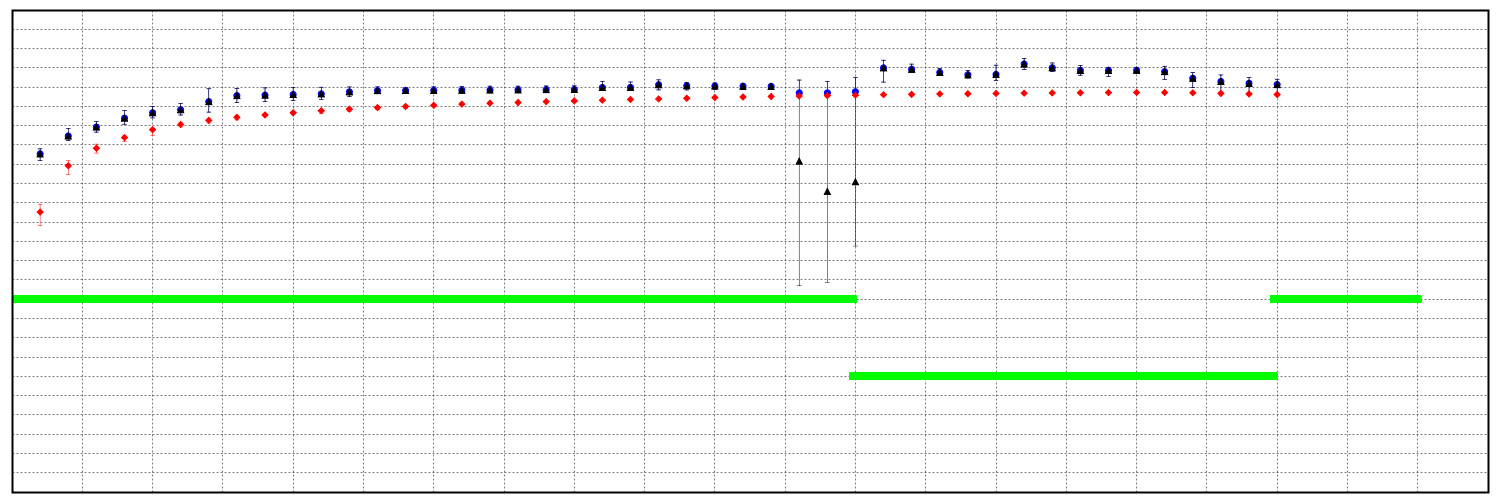}} & \parbox[c][3.6cm][c]{4cm}{\includegraphics[width=4cm,height=3.3cm]{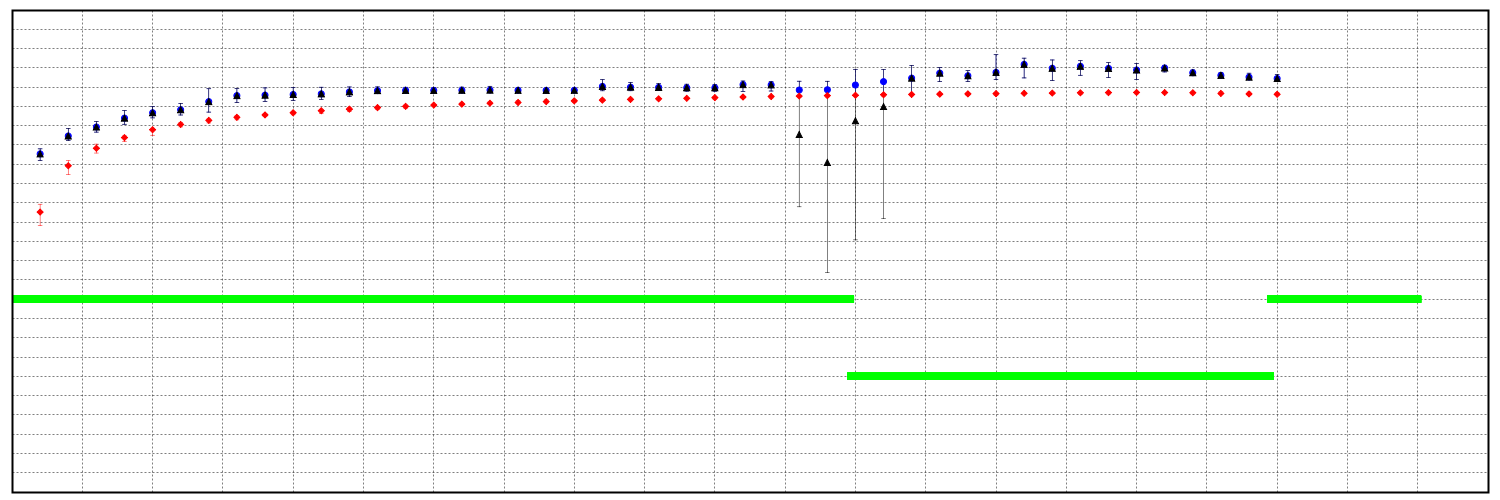}} &
\parbox[c][3.6cm][c]{4cm}{\includegraphics[width=4cm,height=3.3cm]{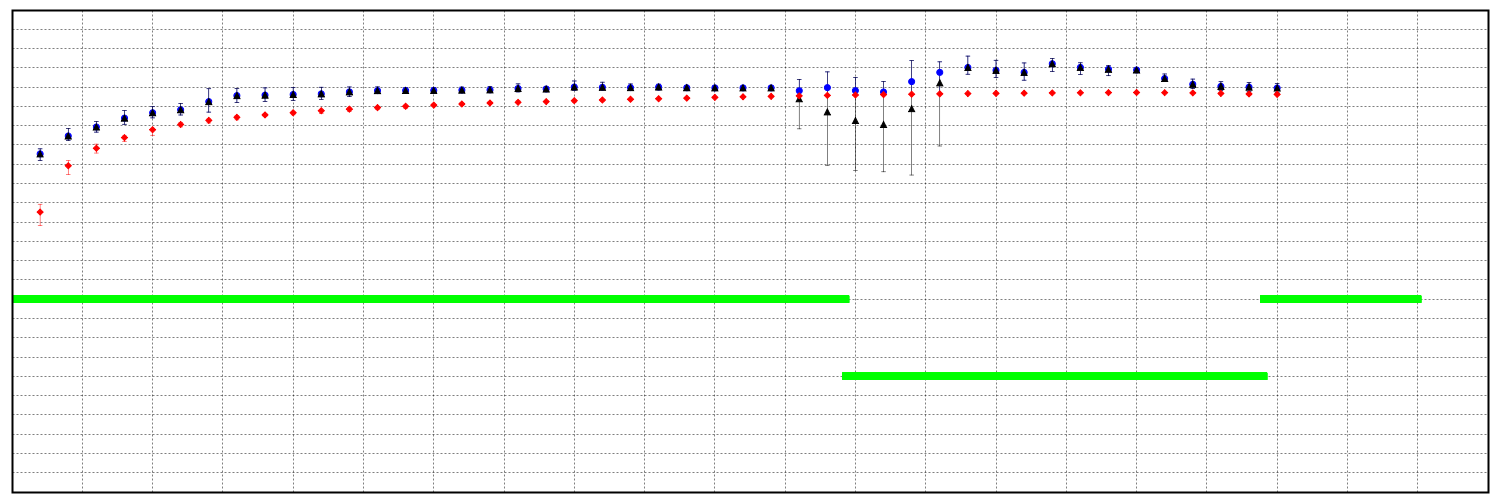}} &
\parbox[c][3.6cm][c]{4cm}{\includegraphics[width=4cm,height=3.3cm]{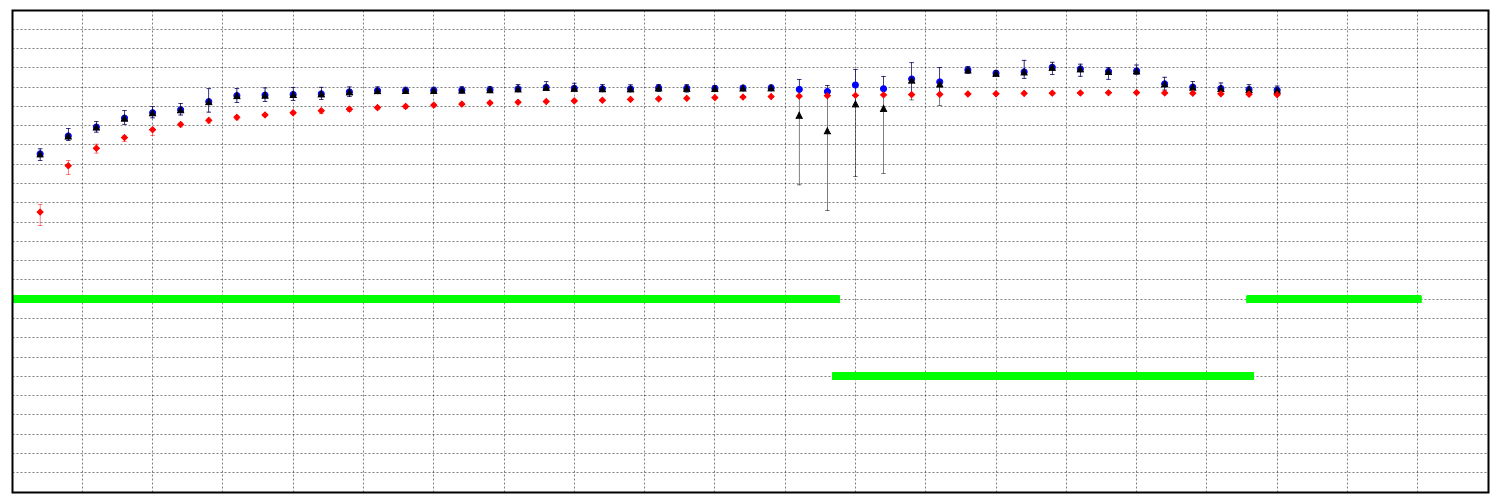}} \\
\hline

\end{tabular}
\label{tab1}
\end{center}
\end{table*}

\end{document}